%% file: main.tex
\documentclass[10pt,twocolumn,a4paper]{article}

\usepackage[final]{cvww}

\usepackage{array}
\usepackage{booktabs}
\usepackage{multirow}
\usepackage[pagebackref,breaklinks,colorlinks,allcolors=cvwwblue]{hyperref}

\def\paperID{10}

\title{Leveraging Intermediate Representations\\ for Better Out-of-Distribution Detection}

\author{Gianluca Guglielmo$^{1}$\and 
Marc Masana$^{1,2}$\vspace{0.2em}\and
{$^{1}$Institute of Visual Computing, TU Graz}\\
{$^{2}$SAL Dependable Embedded Systems, Silicon Austria Labs}\\
{\tt\small \{guglielmo, mmasana\}@tugraz.at}
}

\newcommand{\minisection}[1]{\vspace{0.03in} \noindent {\bf #1}}
\def\rvx{{\mathbf{x}}}
\definecolor{NEAR_OOD}{HTML}{fc8d62}
\definecolor{FAR_OOD}{HTML}{66c2a5}
\definecolor{CORRUPTION}{HTML}{3772FF}
\definecolor{myred}{RGB}{187, 1, 54}
\definecolor{mygreen}{RGB}{61, 120, 56}
\definecolor{othergreen}{rgb}{0.0, 0.5, 0.0}
\definecolor{fgold}{RGB}{212, 175, 55}
\definecolor{ssilver}{RGB}{192, 192, 192}
\definecolor{tbronze}{RGB}{159, 122, 52}
\input{resources.tex}

\begin{document}
\maketitle

\begin{abstract}
In real-world applications, machine learning models must reliably detect Out-of-Distribution (OoD) samples to prevent unsafe decisions. Current OoD detection methods often rely on analyzing the logits or the embeddings of the penultimate layer of a neural network. However, little work has been conducted on the exploitation of the rich information encoded in intermediate layers. To address this, we analyze the discriminative power of intermediate layers and show that they can positively be used for OoD detection. Therefore, we propose to regularize intermediate layers with an energy-based contrastive loss, and by grouping multiple layers in a single aggregated response. We demonstrate that intermediate layer activations improves OoD detection performance by running a comprehensive evaluation across multiple datasets\footnote{Code is available at:  \url{https://github.com/gigug/LIR}}.
\end{abstract}

\section{Introduction}\label{sec:intro}
When a model is exposed to data which does not belong to the distribution it was originally trained on, it is desirable that it can detect it and respond appropriately. Therefore, it is beneficial for a machine learning framework to include an \textit{Out-of-Distribution}~(OoD) detection mechanism, especially in real-world scenarios. Without it, the model might produce unreliable or even dangerous outputs when confronted with data from an unfamiliar distribution, leading to potential failures in critical applications such as autonomous driving, healthcare, or financial systems~\cite{amodei2016concrete, yang2024generalized}. Deep neural networks perform well in many applications but can be overly confident with unseen classes~\cite{nguyen2015deep}. A key feature would be the ability to avoid providing (overconfident) predictions for unknown classes. Implementing this safety mechanism should not interfere with the intended tasks of the model, such as correctly classifying the samples from the \textit{In-Distribution}~(ID) data~\cite{yang2024generalized}. However, achieving a balance between ID performance and OoD detection, presents significant challenges. Furthermore, OoD detection mechanisms ought to perform efficiently, without imposing an excessive computational overhead or diminishing the capacity of the model when performing on the original task. Although recent advances have lead to promising strategies~\cite{lu2024recent, liu2021towards, yang2024generalized}, developing methods that achieve this dual goal remains a pressing challenge for the design of trustworthy, scalable AI systems~\cite{diaz2023connecting}.

Building on these challenges, deep learning models have emerged as a powerful solution, becoming the preferred framework for constructing complex training pipelines~\cite{dong2021survey}. These models address the need for effective OoD detection by leveraging their hierarchical architectures, which enable the learning and encoding of \textit{mid-level features}~\cite{oquab2014learning}, which are representations that bridge low-level patterns such as edges and textures to high-level abstract feature maps such as object parts and semantic categories~\cite{GoodBengCour16}. In computer vision, these features are inherently diverse, capturing the hierarchical nature of the input data. This diversity not only makes them able to generalize to tasks within the original in-distribution but also highly transferable to new, related tasks. This has been shown within transfer learning scenarios~\cite{evci2022head2toe}. 

\figmotivation{Intermediate representations are often more informative than the logits when dealing with OoD detection.}

For OoD detection, most methods rely on penultimate layer embeddings or on the logits of the model~\cite{yang2024generalized}. The potential of leveraging ensembles of intermediate layer embeddings remains under-explored~\cite{lee2018simple}. We argue that mid-level features alone can act as reliable stand-alone OoD-indicators. For instance, inputs with semantically unrelated characteristics compared to the training data may trigger unusual activations in specific layers, serving as an early warning of abnormality.

We pose that specific hidden layers can be effectively isolated and used to enhance OoD detection, performing better than the final layer. However, leveraging these intermediate representations may yield different results depending on the type of shift from ID data -— whether semantic or covariate. Also, this may result in varying effects depending on how close or distant the ID and OoD distributions are.

Building on this insight, we test with an aggregated approach, which leverages hidden-layer information in a layer-agnostic manner. This method avoids relying on specific layers, which enables a more robust and generalized use of the network's intermediate representations for OoD tasks. Further, we also propose an approach that regularizes selected hidden layers through an energy-based contrastive loss, improving OoD detection by leveraging their intermediate representations. The goal is to promote the information encoded in the hidden spaces to be distributed such that OoD detection is more efficient, and without disrupting the ID task performance.

Therefore, our contributions are summarized as:
\begin{itemize}
    \vspace{0.5em}
    \item we establish that the embeddings of hidden layers are valuable for OoD detection,
    \vspace{0.5em}
    \item we introduce a layer-agnostic aggregated (Ag-EBO) approach that leverages intermediate representations,
    \vspace{0.5em}
    \item we propose a modular strategy to enhance robustness by regularizing specific layers (R-EBO). 
    \vspace{0.5em}
\end{itemize}

The article is structured as follows: \cref{sec:related} presents a current overview of the OoD field, while \cref{sec:prelim} introduces the preliminaries needed for the study in \cref{sec:method}, which shows that intermediate layers contain useful information for the OoD detection task. An overview of the proposed ways of exploiting this capabilities, with detailed results is presented in \cref{sec:exps}. Finally, in \cref{sec:discussion} we discuss some limitations of the proposed approaches and the main takeaways.

\section{Related Work}\label{sec:related}
In this paper, we concentrate on two main families of Out-of-Distribution methods: \textit{post-hoc} and \textit{training-based}~\cite{liu2021towards, yang2024generalized}.

\minisection{Post-hoc methods} are applied after the model has been trained and typically involve analyzing its predictions or intermediate representations to identify whether an input is OoD~\cite{guo2017calibration, wei2022mitigating, hendrycks2018baseline}. These methods often focus on computational efficiency and adaptability to pre-trained models, as they avoid retraining~\cite{liu2021towards}.

\minisection{Training-based methods} modify the training process, sometimes completely restructuring the model to accommodate OoD detection~\cite{chang2020generalized, devries2018learning, tack2020csi}. These methods often come at the cost of higher training complexity, and might dilute the efforts to obtain an optimal ID training accuracy~\cite{liu2021towards, masana2018metric}. Additionally, exposure to outliers (real or generated) can be done to improve generalization~\cite{wang2023learning}.

\minisection{Baselines.}
A classic baseline for OoD is considered to be Maximum Softmax Probability (MSP)~\cite{hendrycks2018baseline}, a simple approach that relies on the logit scores to identify OoD samples. However, a major limitation of this approach is the tendency of models to produce overconfident predictions on anomalous data, leading to poor performance~\cite{guo2017calibration}. Temperature scaling~\cite{guo2017calibration} is a simple post-hoc way of tackling the overconfidence issue, where logits are scaled by a temperature $T$, but its results are not optimal~\cite{zhang2024openood}.

\minisection{OoD and intermediate layers.}
Some methods leverage intermediate embeddings within the network. However, most do it to refine the head's detection capabilities, rather than for direct OoD detection. ASH~\cite{djurisic2023extremely} enhances the network's OoD detection capabilities through activation masking of hidden layers. Similarly, ReAct~\cite{sun2021react} proposes to rectify the embeddings of the penultimate layer to reduce overconfidence. However, despite leveraging intermediate embeddings to an extent, the final detection decisions in both methods rely solely on the output logits. Mahalanobis distance-based method (MDSEns)~\cite{lee2018simple} uses features from hidden layers to compute distances from the known distribution. However, this approach relies on the assumption that the class-conditional distributions of hidden layer features are Gaussian, which may not hold true for complex datasets and deep network architectures~\cite{venkataramanan2023gaussian}. Head2Toe~\cite{evci2022head2toe} leverages intermediate representations by training a classifier head on concatenated embeddings from multiple hidden layers to improve generalization during \textit{transfer learning}. This enables the refinement of existing OoD detection techniques through the utilization of hidden layer structures.

\section{Out-of-Distribution Detection}\label{sec:prelim}

\subsection{Problem statement}\label{sec:prob_stat}

In Out-of-Distribution (OoD) detection, the objective is to differentiate between samples generated by the same distribution as the in-distribution dataset, \(\mathcal{D}_{\text{in}}\), and those originating from a different, out-of-distribution dataset, \(\mathcal{D}_{\text{out}}\). Due to the complexity and variance of image-based data, the concept of the amount of \emph{out-of-distributioness} of samples is inherently challenging to define. However, two primary types of distributional shifts are commonly identified~\cite{tian2021exploring}:
\begin{itemize} 
    \item \textbf{Semantic (or Concept) shifts:} they arise when new classes appear at test time. For instance, encountering an image of a dog after the model has been trained on pictures of cats and mice.
    \item \textbf{Covariate shifts:} occur when the style or attributes of samples change within the same class. Examples include image corruptions~\cite{hendrycks2018benchmarking}, such as artifacts, blurs or noise, and domain changes~\cite{hendrycks2021faces, wang2024dissecting}, such as shifting from natural photographs to artistic paintings.
\end{itemize}

\noindent Both semantic and covariate shifts can occur with varying levels of severity depending on the problem, and can also appear entangled within a distribution shift. Given a fixed $\mathcal{D}_{in}$, we refer to \textit{near} and \textit{far} OoD datasets as those that are semantically closer to or further from it, respectively.

Moreover, depending on the OoD detection application, different shifts might be considered within the spectrum that comprises between novelty and anomaly detection~\cite{masana2018metric}. The first relates to distribution shift that might need to be explicitly added to the model, while the second is usually added in a more implicit way, in order to efficiently use the capacity of the model. In this paper, we do not distinguish samples based on the suitability for further learning, but instead aim to analyze these shifts from a perspective of distribution similarity.

\minisection{Terminology.}
Consider a neural network $f(\rvx;\theta)$ with input $\rvx$ and parameters $\theta$, and trained to classify $C$ classes. The architecture of the network is defined as a series of $L$ layers with intermediate functions such that:
\[
y = f(\rvx;\theta) = (f^{\theta_{L}}_{L} \circ f^{\theta_{L-1}}_{L-1} \circ \dots \circ f^{\theta_{1}}_{1})(\rvx),
\]
\noindent where the output $y$ is a vector of $C$ logits representing the unnormalized prediction over the classes. Therefore, the intermediate representations or embeddings of a given layer $l$ are defined as:
\[
a_{l} = (f_{l} \circ \dots \circ f_{1})(\rvx).
\]
To determine whether an input $\rvx$ belongs to $\mathcal{D}_{\text{in}}$ or $\mathcal{D}_{\text{out}}$, a score function $\mathcal{S}(\rvx)$, is usually derived from the neural network. This score reflects the confidence of the model in the input belonging to the expected in-distribution. A threshold $T$ is applied to classify the input such that: 
\[
g(\rvx) = \ \left\{
\begin{array}{ll}
\rvx\in\mathcal{D}_{\text{in}} & \text{if } \mathcal{S}(\rvx) \geq T \\
\rvx\in\mathcal{D}_{\text{out}} & \text{if } \mathcal{S}(\rvx) < T.
\end{array}
\right.
\]
The threshold can be adjusted depending on the desired balance between sensitivity and specificity for OoD detection. 

\minisection{Metrics.} In order to evaluate the strength of a method, two essentials metrics are \mbox{AUROC}, the Area Under the Receiver Operating Characteristic (the higher the better) and \mbox{FPR@TPR95}, the False Positive Rate when the True Positive Rate is 95\% (the lower the better).

\subsection{Energy-based out-of-distribution detection.}
Energy-based models~\cite{lecun2006tutorial} have demonstrated to be effective as post-hoc OoD detectors. The \textit{free energy function} $E(\rvx; \mathbf{f})$ is defined as:
\begin{equation}\label{eq:1}
    E(\rvx; \mathbf{f}) = -T \log \sum_{c=1}^{C} e^{f^{c}(\rvx)/T} \,,
\end{equation}
where $T$ is the temperature, for temperature scaling~\cite{guo2017calibration}. When $T\!=\!1$, it simplifies to the negative log of the denominator of the softmax function, which represents the normalization factor in the softmax computation. In this case, the energy function effectively captures the aggregate contribution of all logits, weighted by their exponential, to produce a measure of confidence over the entire output distribution.  The Energy-Based OoD (EBO)~\cite{liu2020energy} detection approach uses the free energy associated to each input to determine whether it is ID or OoD, where the higher the energy is, the more likely the sample is OoD. JEM~\cite{grathwohl2020jem} is another energy-based approach that improves the calibration (the mismatch between accuracy and confidence) of the model.

\section{OoD with Intermediate Layers}\label{sec:method}

\subsection{Motivation}
As data moves through the trained layers of the network, the represented features become more complex, from edges and simple texture patterns to higher-level representations or combinations of intermediate features~\cite{oquab2014learning}. Our assumption is that the use of these intermediate representations can improve out-of-distribution detection. Therefore, we take EBO~\cite{liu2020energy} as a starting point and analyze how discriminative the different layers of the model are for OoD detection. To quantify the capacity of hidden layers in the OoD task, we introduce a hypothetical method called \textit{Best Hidden Layer} (BHL), which utilizes an oracle to identify the optimal hidden layer for OoD detection. Therefore, since it requires access to the distribution ground truth, it is proposed as an a-posteriori analysis strategy.

Following classic setups, we train a classification model on $\mathcal{D}_{\text{in}}$, using the standard \textit{cross-entropy} loss $\mathcal{L}_{\text{CE}}$. Then, we evaluate on test data from both $\mathcal{D}_{\text{in}}$ and $\mathcal{D}_{\text{out}}$, extracting the embeddings from the intermediate layers for each sample.
Here, the free energy from~\cref{eq:1} is a natural candidate to use on the logits. However, the function can also take the embeddings $\mathbf{a}_{l}$ from any other layer $l$. Thus, we propose to extract the energy score:

\begin{equation}\label{eq:2}
    E_{l}(\rvx) = -T \log \sum_{i} e^{a_{l}^{i}(\rvx)/T} \,,
\end{equation}
where the unit indices $i$ correspond to the output of the \mbox{\textit{l}-th} layer. 

We extract and analyze the energy of each layer, regardless of its type, such as convolutional, batch normalization, or fully connected. We observe that certain intermediate layers consistently outperform the network logits from the original EBO approach. This effect is shown in \Cref{fig:cifar10} for semantic shift, which presents the AUROC scores evaluated across all layers of a ResNet18~\cite{he2016deep} for \textsc{CIFAR-10}~\cite{krizhevsky2009learning} as $\mathcal{D}_{\text{in}}$ and near and far OoD as $\mathcal{D}_{\text{out}}$. Some high-performing layers exhibit unexpected behavior by assigning lower energy values to $\mathcal{D}_{out}$ samples instead of $\mathcal{D}_{in}$ samples. This leads to two possibilities: assigning OoD to lower energy samples or to higher energy samples. Among the two, the ``correct'' possibility is reflected in the reported results. 

\figcifar{AUROC scores for OoD detection for each intermediate layer of ResNet18 are presented. The network is pretrained on \textsc{CIFAR-10} ($\mathcal{D}_{\text{in}}$) and evaluated against the corresponding $\mathcal{D}_{out}^{near}$ and $\mathcal{D}_{out}^{far}$ datasets. Results are averaged across datasets in both categories.}

Covariate shift OoD detection also shows significant improvement when considering intermediate layers rather than relying solely on the network’s output logits. To test it, we look at the performance of different layers when the OoD represents the in-distribution shifted by different corruptions (\textsc{CIFAR-10-C}~\cite{hendrycks2018benchmarking}, see \cref{sec:implementation}). \Cref{fig:cifar10_corruptions} shows that throughout the depth of the network, several layers outperform yet again the head. Initial layers, which provide low-level features such as edges or local histogram projections, seem to be good candidates for OoD detection when covariate shift is present, since it represents a transformation on the in-distribution.

Despite the clear benefits from using some of the layers, determining which one to use for OoD detection under different shifts is still challenging due to different \(\mathcal{D}_{out}\) distributions or modes having a tendency to elicit the strongest responses in different layers. This variability means that no single layer is universally optimal for detecting all types of OoD inputs effectively.

It must be noted that, on average for semantic shifts, the optimal layers are observed to reside more towards the later layers of the network (see \cref{fig:cifar10}). However, this is not enough to identify a good one-fits-all layer, or to find a straight-forward selection criteria.
We try to circumvent this issue by proposing two strategies to leverage the information from the intermediate layer representation spaces:
\begin{itemize}
    \item aggregating all intermediate responses into a single unified response (described in~\cref{sec:aggregation});
    \item strictly regularize selected layers to enforce generalization over different distributions (described in~\cref{sec:regularization}).
\end{itemize}

\figcifarcorruptions{AUROC scores for each intermediate layer of ResNet18 pretrained on \textsc{CIFAR-10} as $\mathcal{D}_{in}$ and evaluated against different corruptions (\textsc{CIFAR-10-C}). Results are averaged over all corruption types and seeds.}

\begin{table*}[t]
    \centering
    \begin{tabular}{c@{\hskip 0.2in}c@{\hskip 0.2in}c@{\hskip 0.2in}c@{\hskip 0.2in}c}
     \toprule
     &\textbf{CIFAR-10}~\cite{krizhevsky2009learning} & \textbf{CIFAR-100}~\cite{krizhevsky2009learning} & \textbf{ImageNet200}~\cite{deng2009imagenet} & \textbf{ImageNet}~\cite{deng2009imagenet}
     \\
     \midrule
    \multirow{1}{*}{\textbf{Architecture}} & \textsc{ResNet18}~\cite{nguyen2015deep} & \textsc{ResNet18} & \textsc{ResNet18} & \textsc{ResNet50}~\cite{nguyen2015deep} \\
    \midrule
    \textbf{Input Size} & $32\!\times\!32\!\times\!3$ & $32\!\times\!32\!\times\!3$ & $224\!\times\!224\!\times\!3$ & $224\!\times\!224\!\times\!3$ \\
    \midrule
    \multirow{2}{*}{\textbf{Near-OoD}} 
    & \textsc{CIFAR-100} & \textsc{CIFAR-10} & \textsc{SSB-Hard}~\cite{zhang2024openood} & \textsc{SSB-Hard} \\ 
    & \textsc{TinyImageNet}~\cite{deng2009imagenet} & \textsc{TinyImageNet} & \textsc{Ninco}~\cite{bitterwolf2023ninco} & \textsc{Ninco} \\
     \midrule
    \multirow{4}{*}{\textbf{Far-OoD}} 
    & \textsc{Texture}~\cite{cimpoi14describing} & \textsc{Texture} & \textsc{Texture} & \textsc{Texture} \\
    & \textsc{MNIST}~\cite{deng2012mnist} & \textsc{MNIST} & \textsc{iNaturalist}~\cite{DBLP:journals/corr/HornASSAPB17} & \textsc{iNaturalist} \\
    & \textsc{SVHN}~\cite{netzer2011reading} & \textsc{SVHN} & \textsc{OpenImageO}~\cite{Wang2022ViMOW} & \textsc{OpenImageO} \\
    & \textsc{Places365}~\cite{Zhou2018PlacesA1} & \textsc{Places365} & - & - \\
     \midrule
    \textbf{Corruptions} & \textsc{CIFAR-10-C}~\cite{hendrycks2018benchmarking} & - & - & - \\
    \bottomrule
    \end{tabular}
    \caption{\label{tab:architectures}Setup description for each ID dataset.}
\end{table*}

\begin{table*}[t]
    \centering
    \resizebox{\linewidth}{!}{
    \begin{tabular}{l@{\hskip 0.1in}cca@{\hskip 0.3in}cccca}
        \toprule
        & \textbf{CIFAR-100} & \textbf{TIN} & \textbf{Near OoD} & \textbf{MNIST} & \textbf{Places365} & \textbf{SVHN} & \textbf{Texture} & \textbf{Far OoD} \\
        \midrule
        EBO~\cite{liu2020energy} & \underline{86.36} & \underline{88.80} & \underline{87.58} & 94.32 & \underline{89.25} & 91.79 & 89.47 & \underline{91.21} \\
        \midrule
        BHL & \textbf{88.23} & \textbf{92.26} & \textbf{90.25} & \textbf{99.89} & \textbf{92.13} & \textbf{98.46} & \textbf{93.5} & \textbf{96.00} \\
        \midrule MDSEns~\cite{lee2018simple} & 61.29 & 59.57 & 60.43 & \underline{99.17} & 66.56 & 77.40 & 52.47 & 73.90 \\
        \midrule
        Ag-EBO w/ MD &66.03&67.03&66.53&99.05&63.73&\underline{94.25}&\underline{93.32}&87.59 \\
        Ag-EBO w/ KNN & 83.69 & 86.5 & 85.09 & 92.81 & 86.01 & 89.64 & 87.46 & 88.98 \\
        Ag-EBO w/ VAE & 80.42 & 83.11 & 81.77 & 89.31 & 83.27 & 88.25 & 84.13 & 86.24\\
        \bottomrule
    \end{tabular}
    }
    \caption{AUROC scores of MDSEns, EBO, BHL and three aggregation methods with \textsc{CIFAR-10} as $\mathcal{D}_{\text{in}}$, averaged over 3 runs.}
    \label{tab:cifar10}
\end{table*}

\subsection{Energy aggregation (Ag-EBO)}\label{sec:aggregation}
To develop a fully layer-agnostic post-hoc method that leverages all the potential from intermediate embeddings, we propose to aggregate the energy values extracted from all \(L\) layers simultaneously. Thus, for each input $\rvx$, we construct a vector of energies:
\[
\mathbf{E}(\rvx) = (E_{1}(\rvx), \dots, E_{L}(\rvx)),
\]
which groups the energy contributions of each layer into a unified representation.
The dimension of this vector is significantly smaller than the total hidden dimension of the network, making it scalable and suitable for use with most common OoD methods. However, for the intermediate layer to be considered, it is desirable that it offers better results than just relying on the logits or on the embeddings from the penultimate layer.

We tested with some straightforward approaches from literature, presented in the next paragraphs. Two of the following three methods need a reference for the ID data, therefore we use the set of energies \mbox{$\tilde{E}=E_{\text{in}}^{\text{train}} = \{\mathbf{E}(\rvx)\,|\,\rvx\!\in\!\mathcal{D}_{\text{in}}^{\text{train}}\}$}, extracted from $\mathcal{D}_{\text{in}}^{\text{train}}$.

\vspace{0.3em}
\minisection{Mahalanobis distance.}
The score $\mathcal{S}_{\text{MD}}(\rvx)$ depends on the Mahalanobis distance~\cite{lee2018simple} of $\mathbf{E(\rvx)}$:
\[
\mathcal{S}_{\text{MD}}(\rvx) = \min_{\boldsymbol{\mu}_c \in \tilde{E}} \sqrt{\left( \mathbf{E}(\rvx) - \boldsymbol{\mu}_c \right)^\top \mathbf{\Sigma}_c^{-1} \left( \mathbf{E}(\rvx) - \boldsymbol{\mu}_c \right)},
\]
where $\boldsymbol{\mu}_c$ and $\mathbf{\Sigma}_c$ are the mean vector and covariance matrix of the energy vectors for class $c$ in $\mathcal{D}_{\text{in}}^{\text{train}}$, respectively.

\vspace{0.3em}
\minisection{K-nearest neighbor.}  
The score $\mathcal{S}_{\text{KNN}}(\rvx)$ is based on the distance of $\mathbf{E}(\rvx)$ to its $K$ nearest neighbors~\cite{sun2022out}:
\[
\mathcal{S}_{\text{KNN}}(\rvx) = \frac{1}{K} \sum_{i=1}^K \Big\| \mathbf{E}(\rvx) - \mathbf{E}_i \Big\|_2,
\]
where $\{\mathbf{E}_1, \dots, \mathbf{E}_K\} \subset \tilde{E}$ are the $K$ nearest neighbors of $\mathbf{E}(\rvx)$ in the in-distribution training set, measured using the Euclidean distance.

\vspace{0.3em}
\minisection{Reconstruction Error.}  
The score $\mathcal{S}_{\text{VAE}}(\rvx)$ is computed as the reconstruction error of $\mathbf{E}(\rvx)$ using a small Variational Autoencoder~\cite{kingma2022autoencoding}: 
\[
\mathcal{S}_{\text{VAE}}(\rvx) = \left\| \mathbf{E}(\rvx) - \mathbf{\hat{E}}(\rvx) \right\|_2,
\]
where $\mathbf{\hat{E}}(\rvx)$ is the reconstruction of $\mathbf{E}(\rvx)$. Higher reconstruction error indicates that the input is likely to be out-of-distribution.

\subsection{Energy regularization (R-EBO)}\label{sec:regularization}
Regularizing intermediate layers directly provides an effective approach to addressing the intermediate layer selection problem. Ideally, by enforcing a strong energy-based discriminative behavior within the hidden layers, we promote their reliability, allowing them to be used confidently without additional selection mechanisms.

EBO~\cite{liu2020energy} introduces an energy-bounded learning loss $\mathcal{L}_{\text{energy}}$ to push the network to assign low energy values to ID samples (and viceversa for OoD). Since their approach operates at the logits level, this loss is applied exclusively to the model's head.
In contrast, our proposed strategy extends the scope of this loss by applying it to each hidden convolutional layer during training, computing and back-propagating all the losses simultaneously. Given an ID dataset $\mathcal{D}_{\text{in}}^{\text{train}}$ and an OoD seen dataset $\mathcal{D}_{\text{out}}^{\text{train}}$ (for outlier exposure), the energy regularization loss for the \textit{l}-th hidden layer is defined as:
\begin{equation}\label{eq:3}
    \begin{aligned}
        \mathcal{L}_{\text{energy},l} &= \mathbb{E}_{\rvx_{\text{in}} \sim \mathcal{D}_{\text{in}}^{\text{train}}}[\max(0, E_{l}(\rvx_{\text{in}}) - m_{\text{in}})]^2 \\ &+ \mathbb{E}_{\rvx_{\text{out}} \sim \mathcal{D}_{\text{out}}^{\text{train}}}[\max(0, m_{\text{out}} - E_{l}(\rvx_{\text{out}})]^2\,,
    \end{aligned}
\end{equation}
where $m_{\text{in}}$ and $m_{\text{out}}$ are two margins, serving as the upper bound for the energy of the ID data and the lower bound for the energy of the seen OoD data, respectively. We define the total loss as:
\begin{equation}\label{eq:4}
        \mathcal{L}_{\text{R-EBO}} = \sum_{l=1}^{L} \mathcal{L}_{\text{energy},l}\,,
\end{equation}
where the same constant margin values for $m_{\text{in}}$ and $m_{\text{out}}$ are used across all layers, although each can be explored independently. In the original EBO paper~\cite{liu2020energy}, \mbox{$\mathcal{L}_{\text{EBO}} = \mathcal{L}_{\text{energy},E_{L}}$}, where $L$ is the last layer of the network. Furthermore, the decision to reduce the free ID energy and increase the OoD energy in intermediate layers is a design choice. Alternative regularization strategies can also be considered.

\section{Experimental results}\label{sec:exps}
\subsection{Implementation details}\label{sec:implementation}

\minisection{Datasets.} The datasets used in this study were selected based on the guidelines of the OpenOoD benchmark~\cite{zhang2024openood}, which offers a comprehensive and well-documented collection of state-of-the-art (SoTA) methods across various OoD scenarios. Also, the results presented here have been extracted from its continuously updated report, to ensure alignment with the latest developments in the field. For each $\mathcal{D}_{in}$, the OpenOoD benchmark defines a set of semantically \textit{near} and \textit{far} OoD datasets from it~\Cref{tab:architectures}. Additionally, we tested the response to covariate shift from \textsc{CIFAR-10} with the corruptions dataset \mbox{\textsc{CIFAR-10-C}}~\cite{hendrycks2018benchmarking}.
This is a dataset consisting of corrupted versions of \textsc{CIFAR-10} images, which serves as a common benchmark for evaluating robustness to covariate shifts. It includes a variety of corruption types, such as noise, blur, and weather distortions, applied at varying levels of severity. 

\minisection{Architectures.} To keep the consistency with OpenOoD evaluations, the main results have been calculated using the same architectures used in the benchmark, shown in \cref{tab:architectures}. We also evaluate on a non-residual based convolutional neural network, \textsc{EfficientNet-B7}, for which we select convolutional, fully-connected, batch normalization and average pooling layers. Finally, following recent trends in machine learning, we evaluate ViT-B-16~\cite{dosovitskiy2021an}, a transformer-based~\cite{waswani2017attention} architecture. ViTs utilize multi-head self-attention layers, and their feed-forward sub-layers consist of fully-connected layers. Our experiments focus on the selection of these fully-connected layers for BHL.

\minisection{Training.} 
OpenOoD provides three pretrained ResNet18 checkpoints for \textsc{CIFAR-10}, \textsc{CIFAR-100}, and \textsc{ImageNet200} as $\mathcal{D}_{in}$, and a single pretrained ResNet50 checkpoint for \textsc{ImageNet}, all trained using standard SoftMax loss.
Additionally, we trained 3 checkpoints for both \textsc{CIFAR-10} and \textsc{CIFAR-100} as $\mathcal{D}_{in}$ using the hidden regularization approach.

\subsection{Analysis of OoD with intermediate layers}
\label{sec:analysis}

In \Cref{tab:cifar10}, \textsc{CIFAR-10} is selected as $\mathcal{D}_{\text{in}}$. EBO refers to the standard energy-based OoD detection mechanism applied directly at the logit level, while BHL shows the energy-based OoD detection using the best performing hidden layer. The results presented for BHL are averaged across the best hidden layer identified in each run, which tends to slightly vary between runs. For every $\mathcal{D}_{\text{out}}$ the results are strongly improved by (at least) one hidden layer's response. It is important to mention that the results presented only consider the internal behavior of the network, while an algorithm which correctly weighs the importance of a layer for OoD detection would also take the head of the model into consideration, potentially merging the best results of the two rows.

\minisection{Energy aggregation.}
The last rows of \Cref{tab:cifar10} present the results of the aggregation methods (Ag-EBO) proposed in \Cref{sec:method}. The row above displays the results of \mbox{MDSEns}~\cite{lee2018simple}, taken from the OpenOoD benchmark~\cite{zhang2024openood}.

\noindent Each of our proposed aggregation methods achieves higher AUROC compared to MDSEns~\cite{lee2018simple}, an ensemble method that exploits Mahalanobis distance on hidden layers. The lower results for MDSEns might be related to their assumption of class-conditional distribution of the hidden features being Gaussian. $\mathcal{D}_{\text{out}}^{\text{far}}$ datasets, such as \textsc{MNIST}, \textsc{SVHN}, and \textsc{Texture}, demonstrate improved performance with the KNN aggregation approach compared to EBO. However, none of these methods are robust enough on average to consistently outperform relying exclusively on the head logits. This indicates that the layer-selection problem remains unsolved and cannot yet be effectively simplified into an aggregation mechanism.

\minisection{Energy regularization.}
\Cref{tab:fulloutseen} presents the results of regularization against other SoTA methods that exploit $\mathcal{D}_{\text{out}}^{\text{seen}}$. The margin values are set to \mbox{$m_{in}\!\!=\!\!-25$} and \mbox{$m_{out}\!\!=\!\!-7$}, following the original EBO setup~\cite{liu2020energy}. In order to test the trade-off in hidden layer regularization compared to a completely post-hoc hidden layer analysis, we selected \textsc{CIFAR-10} and \textsc{CIFAR-100} as $\mathcal{D}_{\text{in}}$ and \textsc{ImageNet} as $\mathcal{D}_{\text{out}}^{seen}$. We then trained 5 runs using only $\mathcal{L}_{CE}$, and 5 runs using $\mathcal{L}_{CE} + \mathcal{L}_{\text{R-EBO}}$. We opted not to use the checkpoints given by OpenOoD to guarantee a fair comparison between the two losses. Therefore, the EBO results are not comparable with the ones presented in other tables, and are marked with $(*)$ accordingly. Moreover, only results related to Far-OoD are presented, since Near-OoD includes \textsc{TIN}, which is based on \textsc{ImageNet}.

\begin{table}[t]
    \centering
    \resizebox{\linewidth}{!}{
    \begin{tabular}{l@{\hskip 0.2in}cc@{\hskip 0.2in}cc} 
        \toprule
         & \multicolumn{2}{l}{\textbf{CIFAR-10}} & \multicolumn{2}{l}{\textbf{CIFAR-100}} \\
         & Far & ID Acc. & Far & ID Acc. \\
        \midrule
        EBO~\cite{liu2020energy}* & 84.86 & \textbf{82.33} & 67.86 & \textbf{54.83} \\
        \midrule
        BHL* & 90.42 & \textbf{82.33} & 86.98 & \textbf{54.83} \\
        R-EBO* & \textbf{98.48} & 78.2 & \textbf{94.06} & 50.05 \\
        \bottomrule
    \end{tabular}
    }
    \caption{AUROC scores of EBO, BHL and R-EBO with \textsc{CIFAR-10} as $\mathcal{D}_{\text{in}}$, averaged over multiple runs. EBO and BHL exploit identical checkpoints, retrained (*) for direct comparability with R-EBO.}
    \label{tab:fulloutseen}
\end{table}

\begin{table}[t]
    \centering
    \setlength{\tabcolsep}{6pt}
    \renewcommand{\arraystretch}{1.2}
    \begin{tabular}{l@{\hskip 0.3in}ccc}
        \toprule
        \textbf{Dataset} & \textbf{EBO}~\cite{liu2020energy} & \textbf{BHL} & \textbf{R-EBO} \\
        \midrule
        \textsc{Brightness}          & 56.51 & \textbf{82.98} & \underline{79.8} \\
        \textsc{Contrast}            & 92.39 & \textbf{99.92} & \underline{96.99} \\
        \textsc{Defocus Blur}        & 84.65 & \textbf{97.26} & \underline{85.44} \\
        \textsc{Elastic}             & 73.24 & \textbf{87.36} & \underline{85.11} \\
        \textsc{Fog}                 & 71.3  & \textbf{96.7}  & \underline{94.38} \\
        \textsc{Frost}               & 76.83 & \textbf{91.4}  & \underline{91.08} \\
        \textsc{Gaussian Blur}       & \underline{89.8}  & \textbf{98.86} & 75.96 \\
        \textsc{Gaussian Noise}      & 84.39 & \textbf{99.65} & \underline{99.16} \\
        \textsc{Glass Blur}          & 85.77 & \underline{88.45} & \textbf{98.13} \\
        \textsc{Impulse Noise}       & 89.04 & \textbf{99.98} & \underline{97.71} \\
        \textsc{JPEG}                & \underline{73.33} & \textbf{87.87} & 61.47 \\
        \textsc{Motion Blur}         & \underline{75.78} & \textbf{93.51} & 72.77 \\
        \textsc{Pixelate}            & 80.02 & \underline{94.17} & \textbf{99.66} \\
        \textsc{Saturate}            & 57.47 & \textbf{90.97} & \underline{81.8} \\
        \textsc{Shot Noise}          & 84.93 & \textbf{99.38} & \underline{98.78} \\
        \textsc{Snow}                & 71.85 & \textbf{89.11} & \underline{74.95} \\
        \textsc{Spatter}             & 71.0  & \textbf{90.33} & \underline{83.69} \\
        \textsc{Speckle Noise}       & 85.29 & \textbf{98.96} & \underline{98.66} \\
        \textsc{Zoom Blur}           & \underline{79.36} & \textbf{96.61} & 66.11 \\
        \bottomrule
    \end{tabular}
    \caption{AUROC scores of EBO, BHL, and R-EBO with CIFAR-10 as $\mathcal{D}_{in}$ against corruption datasets.}
    \label{tab:corruption}
\end{table}
    
\begin{table*}[ht]
    \centering
    \begin{tabular}{cccccccccc} 
        \toprule
        & & \multicolumn{2}{c}{\textbf{CIFAR-10}} & \multicolumn{2}{c}{\textbf{CIFAR-100}} & \multicolumn{2}{c}{\textbf{ImageNet-200}} & \multicolumn{2}{c}{\textbf{ImageNet-1K}} \\
        & & Near & Far & Near & Far & Near & Far & Near & Far\\
        \midrule
        \multirow{2}{*}{ResNet18/50} & EBO & 87.58 & 91.21 & \textbf{80.91} & 79.77 & 82.50 & \textbf{90.86} & 75.89 & 89.47 \\
        & BHL & \textbf{90.25} & \textbf{96.00} & 71.57 & \textbf{86.08} & \textbf{86.72} & 76.13 & \textbf{79.04} & \textbf{89.75}\\
        \midrule
        \multirow{2}{*}{EfficientNet-B7} & EBO & \textbf{97.39} & 98.91 & \textbf{87.46} & 86.91 & 75.02 & 86.53 & 65.16 & 81.65 \\
        & BHL & 87.43 & \textbf{99.74} & 84.21 & \textbf{99.80} & \textbf{78.83} & \textbf{93.02} & \textbf{85.24} & \textbf{94.49} \\
        \midrule
        \multirow{2}{*}{ViT-B-16} & EBO & \textbf{90.91} & 93.9 & \textbf{88.81} & 87.23 & \textbf{69.72} & \textbf{83.49} & 62.93 & 78.71 \\
        & BHL & 79.38 & \textbf{96.14} & 81.38 & \textbf{97.98} & 62.19 & 81.40 & \textbf{74.06} & \textbf{88.43} \\
        
        \bottomrule
    \end{tabular}
    \caption{EBO and BHL compared on different models.}
    \label{tab:otherarchs}
\end{table*}

As expected, the regularization of intermediate layers strongly improves the OoD detection capabilities of the model on both cases. However, this comes at the cost of a slight decrease in ID accuracy, due to the additional $\mathcal{L}_{\text{R-EBO}}$ loss.

\minisection{Covariate shift.} \Cref{tab:corruption} presents the detailed OoD results of \textsc{CIFAR-10} against every corruption type present in \textsc{CIFAR-10-C}. As with the semantic shift, we observe that covariate shift is better identified by the hidden layers rather than by the final logits.
\Cref{tab:corruption} also presents R-EBO results under covariate shift conditions, evaluated using the same checkpoints from \Cref{tab:fulloutseen}. The findings suggest that regularizing layers with a semantically distinct $\mathcal{D}_{\text{out}}^{seen}$ does not consistently enhance the identification of covariate shift.

\subsection{Analysis on different architectures}
\Cref{tab:otherarchs} presents the complete results for EBO and BHL, averaged over multiple runs, using \textsc{ResNet18/50}, \textsc{EfficientNet-B7}, and \textsc{ViT-B-16} as backbones. The findings are consistent with earlier observations: BHL improves performance in most setups, except for certain Near OoD cases.

\section{Discussion and Limitations}\label{sec:discussion}

Our findings show that intermediate representations are capable of discriminating out-of-distribution samples better than the logits. Both semantic, in the form of unseen classes, and covariate shift, in the form of image corruptions, are strongly captured by intermediate layers. However, a robust selection criterion for which layer to use is still an open question, since the proposed aggregation method underperforms compared to simpler logit-based alternatives.

Regularization of the intermediate layer's energies improves the results even further, albeit with a trade-off in ID accuracy. We suspect that the influence of $\mathcal{D}_{\text{out}}^{\text{seen}}$ leads to sub-optimal filters for the discrimination of ID classes, thus motivating further research involving regularization which exploits $\mathcal{D}_{\text{in}}$ only. Additionally, regularization using synthetic generated data~\cite{zhu2023diversified} applied to intermediate layers could also be a promising direction, as it would reduce dependence on specific datasets, promote privacy-preservation, and enhance the generalization.

Finally, the findings on this paper pave the way for real-time optimized out-of-distribution detection, enabling the identification of OoD samples in earlier layers during network propagation. By detecting such samples promptly, the system can flag them and halt further processing, reducing computational overhead and improving efficiency.

\section*{Acknowledgements}
Gianluca Guglielmo acknowledges the support of KAI GmbH and Infineon Technologies Austria.
Marc Masana acknowledges the support by the “University SAL Labs” initiative of Silicon Austria Labs (SAL).

{
    \small
    \bibliographystyle{ieeenat_fullname}
    \bibliography{references}
}

\end{document}

%% file: resources.tex
\newcommand{\figmotivation}[1]{
    \begin{figure}[t]
    \centering
    \includegraphics[width=0.95\linewidth, trim=20 42 29 19, clip]{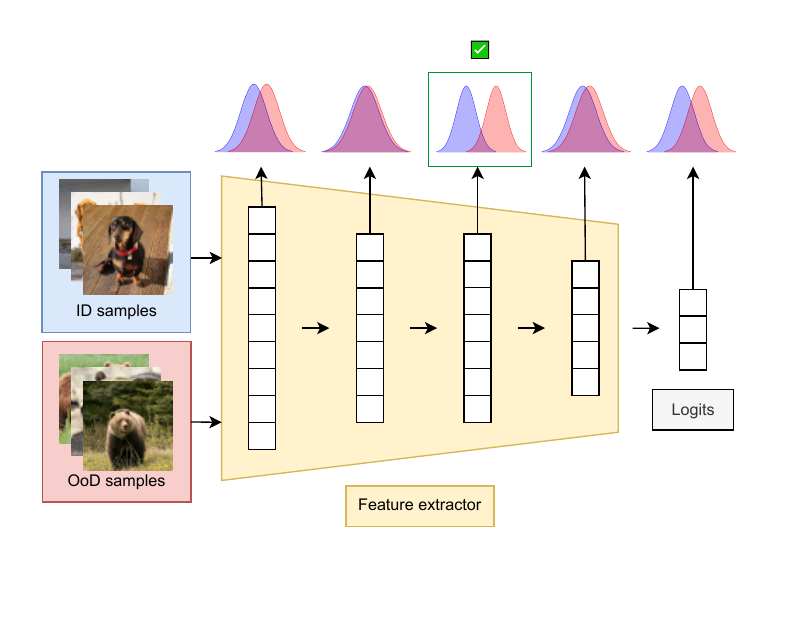}
    \caption{#1}
    \label{fig:explainer}
\end{figure}
}

\newcommand{\figcifar}[1]{
    \begin{figure}[t]
        \centering
        \includegraphics[width=\linewidth, trim=0 0 0 0, clip]{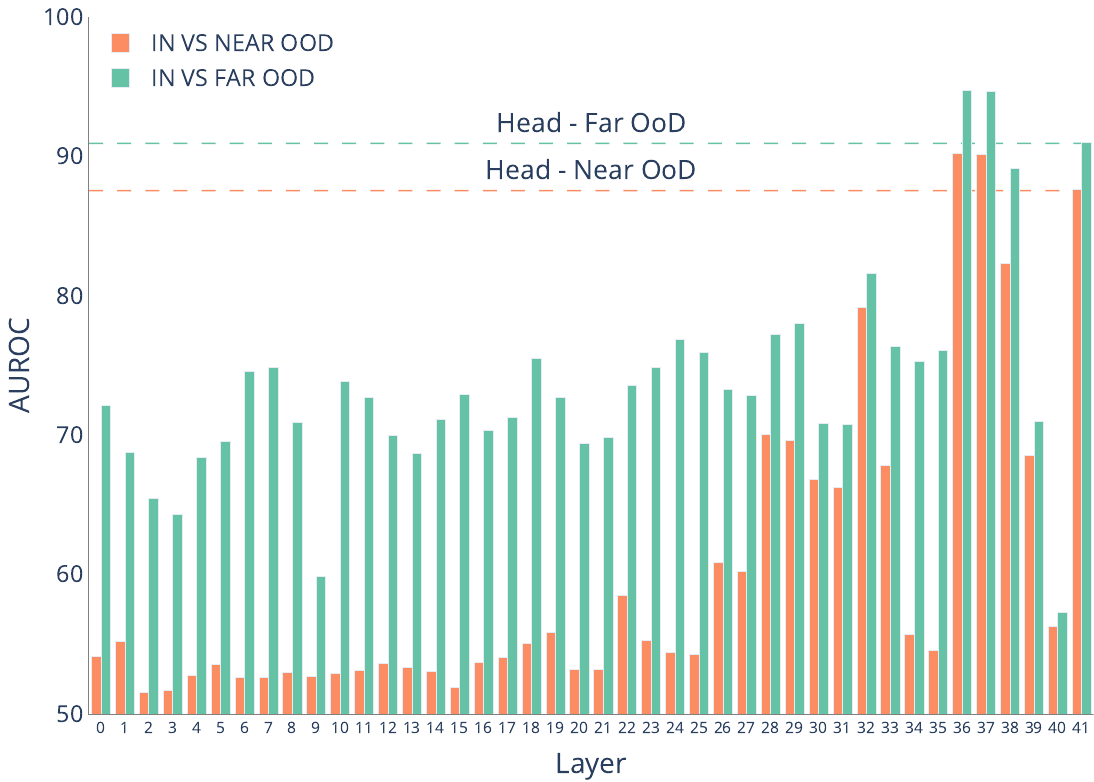}
        \caption{#1}
        \label{fig:cifar10}
    \end{figure}
}

\newcommand{\figcifarcorruptions}[1]{
    \begin{figure}[t]
        \centering
        \includegraphics[width=\linewidth, trim=0 0 0 0, clip]{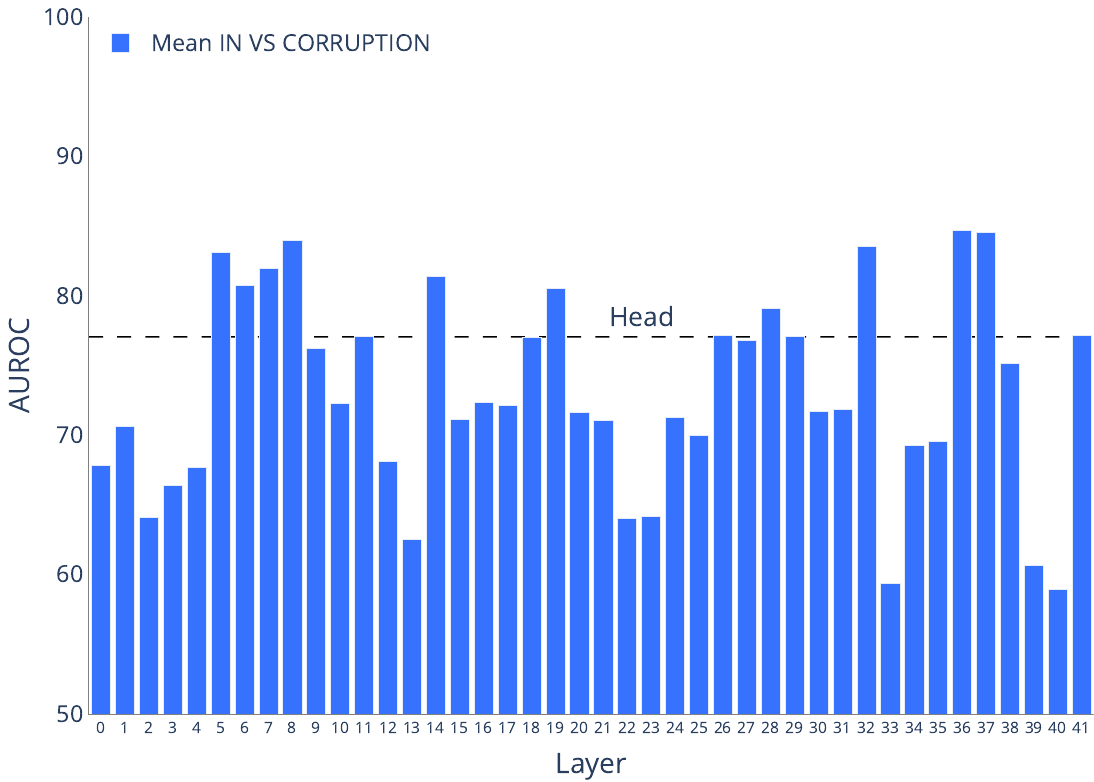}
        \caption{#1}
        \label{fig:cifar10_corruptions}
    \end{figure}
}

\newcommand{\tabledataset}[1]{
    \begin{table}[h]
        \centering
        \small
        \begin{tabular}{ccc}
         \toprule
         \textbf{CIFAR-10} & \textbf{CIFAR-100} & \textbf{ImageNet/ImageNet200} \\
         \midrule
         \rowcolor{NEAR_OOD}
         CIFAR-100 & CIFAR-10 & SSB-Hard \\
         \rowcolor{NEAR_OOD}
         TinyImageNet & TinyImageNet & Ninco \\
         \rowcolor{FAR_OOD}
         Texture & Texture & Texture \\
         \rowcolor{FAR_OOD}
         MNIST & MNIST & iNaturalist\\
         \rowcolor{FAR_OOD}
         SVHN & SVHN & OpenImageO \\
         \rowcolor{FAR_OOD}
         Places365 & Places365 &\cellcolor{white} - \\
         \cellcolor{CORRUPTION} CIFAR-10-C & - & -\\
         \bottomrule
        \end{tabular}
        \caption{\label{tab:datasets}Table of semantic \textcolor{NEAR_OOD}{Near}, semantic \textcolor{FAR_OOD}{Far} and \textcolor{CORRUPTION}{Corruptions} OoD datasets for each ID dataset.}
    \end{table}
}

\newcommand{\tablearchitectures}[1]{
    \begin{table*}[t]
        \centering
        \begin{tabular}{c@{\hskip 0.2in}c@{\hskip 0.2in}c@{\hskip 0.2in}c@{\hskip 0.2in}c}
         \toprule
         &\textbf{CIFAR-10}~\cite{krizhevsky2009learning} & \textbf{CIFAR-100}~\cite{krizhevsky2009learning} & \textbf{ImageNet200}~\cite{deng2009imagenet} & \textbf{ImageNet}~\cite{deng2009imagenet}
         \\
         \midrule
        \multirow{1}{*}{\textbf{Architecture}} & \textsc{ResNet18}~\cite{nguyen2015deep} & \textsc{ResNet18} & \textsc{ResNet18} & \textsc{ResNet50}~\cite{nguyen2015deep} \\
        \midrule
        \textbf{Input Size} & $32\!\times\!32\!\times\!3$ & $32\!\times\!32\!\times\!3$ & $224\!\times\!224\!\times\!3$ & $224\!\times\!224\!\times\!3$ \\
        \midrule
        \multirow{2}{*}{\textbf{Near-OoD}} 
        & \textsc{CIFAR-100} & \textsc{CIFAR-10} & \textsc{SSB-Hard}~\cite{zhang2024openood} & \textsc{SSB-Hard} \\ 
        & \textsc{TinyImageNet}~\cite{deng2009imagenet} & \textsc{TinyImageNet} & \textsc{Ninco}~\cite{bitterwolf2023ninco} & \textsc{Ninco} \\
         \midrule
        \multirow{4}{*}{\textbf{Far-OoD}} 
        & \textsc{Texture}~\cite{cimpoi14describing} & \textsc{Texture} & \textsc{Texture} & \textsc{Texture} \\
        & \textsc{MNIST}~\cite{deng2012mnist} & \textsc{MNIST} & \textsc{iNaturalist}~\cite{DBLP:journals/corr/HornASSAPB17} & \textsc{iNaturalist} \\
        & \textsc{SVHN}~\cite{netzer2011reading} & \textsc{SVHN} & \textsc{OpenImageO}~\cite{Wang2022ViMOW} & \textsc{OpenImageO} \\
        & \textsc{Places365}~\cite{Zhou2018PlacesA1} & \textsc{Places365} & - & - \\
         \midrule
        \textbf{Corruptions} & \textsc{CIFAR-10-C}~\cite{hendrycks2018benchmarking} & - & - & - \\
        \bottomrule
        \end{tabular}
        \caption{\label{tab:architectures}Setup description for each ID dataset.}
    \end{table*}
}

\newcommand{\tablecorruptionold}[1]{
    \begin{table}[t]
        \centering
        \setlength{\tabcolsep}{6pt}
        \renewcommand{\arraystretch}{1.2}
        \begin{tabular}{l@{\hskip 0.5in}cc}
            \toprule
            \textbf{Dataset} & \textbf{EBO}~\cite{liu2020energy} & \textbf{BHL} \\
            \midrule
            \textsc{Brightness}          & 56.51 & \textbf{82.98} \\
            \textsc{Contrast}            & 92.39 & \textbf{99.92} \\
            \textsc{Defocus Blur}        & 84.65 & \textbf{97.26} \\
            \textsc{Elastic}   & 73.24 & \textbf{87.36} \\
            \textsc{Fog}                 & 71.3  & \textbf{96.7}  \\
            \textsc{Frost}               & 76.83 & \textbf{91.4}  \\
            \textsc{Gaussian Blur}       & 89.8  & \textbf{98.86} \\
            \textsc{Gaussian Noise}      & 84.39 & \textbf{99.65} \\
            \textsc{Glass Blur}          & 85.77 & \textbf{88.45} \\
            \textsc{Impulse Noise}       & 89.04 & \textbf{99.98} \\
            \textsc{JPEG}    & 73.33 & \textbf{87.87} \\
            \textsc{Motion Blur}         & 75.78 & \textbf{93.51} \\
            \textsc{Pixelate}            & 80.02 & \textbf{94.17} \\
            \textsc{Saturate}            & 57.47 & \textbf{90.97} \\
            \textsc{Shot Noise}          & 84.93 & \textbf{99.38} \\
            \textsc{Snow}                & 71.85 & \textbf{89.11} \\
            \textsc{Spatter}             & 71.0  & \textbf{90.33} \\
            \textsc{Speckle Noise}       & 85.29 & \textbf{98.96} \\
            \textsc{Zoom Blur}           & 79.36 & \textbf{96.61} \\
            \bottomrule
        \end{tabular}
        \caption{AUROC scores of EBO and BHL with CIFAR-10 as $\mathcal{D}_{in}$ against corruption datasets, averaged over 1 run.}
        \label{tab:corruption}
    \end{table}
}

\newcommand{\tablecorruption}[1]{
    \begin{table}[t]
        \centering
        \setlength{\tabcolsep}{6pt}
        \renewcommand{\arraystretch}{1.2}
        \begin{tabular}{l@{\hskip 0.3in}ccc}
            \toprule
            \textbf{Dataset} & \textbf{EBO}~\cite{liu2020energy} & \textbf{BHL} & \textbf{R-EBO} \\
            \midrule
            \textsc{Brightness}          & 56.51 & \textbf{82.98} & \underline{79.8} \\
            \textsc{Contrast}            & 92.39 & \textbf{99.92} & \underline{96.99} \\
            \textsc{Defocus Blur}        & 84.65 & \textbf{97.26} & \underline{85.44} \\
            \textsc{Elastic}             & 73.24 & \textbf{87.36} & \underline{85.11} \\
            \textsc{Fog}                 & 71.3  & \textbf{96.7}  & \underline{94.38} \\
            \textsc{Frost}               & 76.83 & \textbf{91.4}  & \underline{91.08} \\
            \textsc{Gaussian Blur}       & \underline{89.8}  & \textbf{98.86} & 75.96 \\
            \textsc{Gaussian Noise}      & 84.39 & \textbf{99.65} & \underline{99.16} \\
            \textsc{Glass Blur}          & 85.77 & \underline{88.45} & \textbf{98.13} \\
            \textsc{Impulse Noise}       & 89.04 & \textbf{99.98} & \underline{97.71} \\
            \textsc{JPEG}                & \underline{73.33} & \textbf{87.87} & 61.47 \\
            \textsc{Motion Blur}         & \underline{75.78} & \textbf{93.51} & 72.77 \\
            \textsc{Pixelate}            & 80.02 & \underline{94.17} & \textbf{99.66} \\
            \textsc{Saturate}            & 57.47 & \textbf{90.97} & \underline{81.8} \\
            \textsc{Shot Noise}          & 84.93 & \textbf{99.38} & \underline{98.78} \\
            \textsc{Snow}                & 71.85 & \textbf{89.11} & \underline{74.95} \\
            \textsc{Spatter}             & 71.0  & \textbf{90.33} & \underline{83.69} \\
            \textsc{Speckle Noise}       & 85.29 & \textbf{98.96} & \underline{98.66} \\
            \textsc{Zoom Blur}           & \underline{79.36} & \textbf{96.61} & 66.11 \\
            \bottomrule
        \end{tabular}
        \caption{AUROC scores of EBO, BHL, and R-EBO with CIFAR-10 as $\mathcal{D}_{in}$ against corruption datasets.}
        \label{tab:corruption}
    \end{table}
}

\definecolor{Gray}{gray}{0.92}
\newcolumntype{a}{>{\columncolor{Gray}}c}

\newcommand{\tablecifarten}[1]{
\begin{table*}[t]
        \centering
        \resizebox{\linewidth}{!}{
        \begin{tabular}{l@{\hskip 0.1in}cca@{\hskip 0.3in}cccca}
            \toprule
            & \textbf{CIFAR-100} & \textbf{TIN} & \textbf{Near OoD} & \textbf{MNIST} & \textbf{Places365} & \textbf{SVHN} & \textbf{Texture} & \textbf{Far OoD} \\
            \midrule
            EBO~\cite{liu2020energy} & \underline{86.36} & \underline{88.80} & \underline{87.58} & 94.32 & \underline{89.25} & 91.79 & 89.47 & \underline{91.21} \\
            \midrule
            BHL & \textbf{88.23} & \textbf{92.26} & \textbf{90.25} & \textbf{99.89} & \textbf{92.13} & \textbf{98.46} & \textbf{93.5} & \textbf{96.00} \\
            \midrule MDSEns~\cite{lee2018simple} & 61.29 & 59.57 & 60.43 & \underline{99.17} & 66.56 & 77.40 & 52.47 & 73.90 \\
            \midrule
            Ag-EBO w/ MD &66.03&67.03&66.53&99.05&63.73&\underline{94.25}&\underline{93.32}&87.59 \\
            Ag-EBO w/ KNN & 83.69 & 86.5 & 85.09 & 92.81 & 86.01 & 89.64 & 87.46 & 88.98 \\
            Ag-EBO w/ VAE & 80.42 & 83.11 & 81.77 & 89.31 & 83.27 & 88.25 & 84.13 & 86.24\\
            \bottomrule
        \end{tabular}
        }
        \caption{AUROC scores of MDSEns, EBO, BHL and three aggregation methods with \textsc{CIFAR-10} as $\mathcal{D}_{\text{in}}$, averaged over 3 runs.}
        \label{tab:cifar10}
    \end{table*}
}

\newcommand{\tablefulloutseen}[1]{
    \begin{table}[t]
    \centering
    \resizebox{\linewidth}{!}{
    \begin{tabular}{l@{\hskip 0.2in}cc@{\hskip 0.2in}cc} 
        \toprule
         & \multicolumn{2}{l}{\textbf{CIFAR-10}} & \multicolumn{2}{l}{\textbf{CIFAR-100}} \\
        
         & Far & ID Acc. & Far & ID Acc. \\
        \midrule

    EBO~\cite{liu2020energy}* & 84.86 & \textbf{82.33} & 67.86 & \textbf{54.83} \\

    \midrule
    
    BHL* & 90.42 & \textbf{82.33} & 86.98 & \textbf{54.83} \\
    
    R-EBO* & \textbf{98.48} & 78.2 & \textbf{94.06} & 50.05 \\
    
    \bottomrule
    \end{tabular} 
    }
    \caption{AUROC scores of EBO, BHL and R-EBO with \textsc{CIFAR-10} as $\mathcal{D}_{\text{in}}$, averaged over multiple runs. EBO and BHL exploit identical checkpoints, retrained (*) for direct comparability with R-EBO.}
    \label{tab:fulloutseen}
    \end{table}
}

\newcommand{\tablesub}[1]{
    \begin{table*}[h]
        \begin{subtable}[]{\textwidth}
            \small
            \centering
            \begin{tabular}{l|ccc|ccc|ccc|ccc} 
                \toprule
                 & \multicolumn{3}{c|}{\textbf{CIFAR-10}} & \multicolumn{3}{c|}{\textbf{CIFAR-100}} & \multicolumn{3}{c|}{\textbf{ImageNet200}} & \multicolumn{3}{c}{\textbf{ImageNet}} \\
                 & Near & Far & ID Acc. & Near & Far & ID Acc. & Near & Far & ID Acc. & Near & Far & ID Acc. \\
                \midrule
            \rowcolor{EXTRA_DATA} EBO~\cite{liu2020energy} &&&&&&&&&&& \\
            \midrule
            \rowcolor{EXTRA_DATA} BHL &&&&&&&&&&& \\
            \rowcolor{EXTRA_DATA} R-EBO &&&&&&&&&&& \\
            \bottomrule
            \end{tabular}
            \caption{\textbf{EfficientNet-B7}~\cite{tan2019efficientnet}}
            \label{tab:enb}
        \end{subtable}\vspace{.5cm}
        \begin{subtable}[]{\textwidth}
        \small
        \centering
        \begin{tabular}{l|ccc|ccc|ccc|ccc} 
            \toprule
             & \multicolumn{3}{c|}{\textbf{CIFAR-10}} & \multicolumn{3}{c|}{\textbf{CIFAR-100}} & \multicolumn{3}{c|}{\textbf{ImageNet200}} & \multicolumn{3}{c}{\textbf{ImageNet}} \\
             & Near & Far & ID Acc. & Near & Far & ID Acc. & Near & Far & ID Acc. & Near & Far & ID Acc. \\
            \midrule
        \rowcolor{EXTRA_DATA} EBO~\cite{liu2020energy} &&&&&&&&&&& \\
        \midrule
        \rowcolor{EXTRA_DATA} BHL &&&&&&&&&&& \\
        \rowcolor{EXTRA_DATA} R-EBO &&&&&&&&&&& \\
        \bottomrule
        \end{tabular}
        \caption{\textbf{ViT-B-16}~\cite{dosovitskiy2021an}}
        \label{tab:vit}
        \end{subtable}\vspace{.5cm}
        \caption{Additional networks.}
    \end{table*}
    
}

\newcommand{\tableotherarcs}[1]{
    \begin{table*}[h]
    \centering
    \begin{tabular}{cccccccccc} 
        \toprule
        & & \multicolumn{2}{c}{\textbf{CIFAR-10}} & \multicolumn{2}{c}{\textbf{CIFAR-100}} & \multicolumn{2}{c}{\textbf{ImageNet-200}} & \multicolumn{2}{c}{\textbf{ImageNet-1K}} \\
        & & Near & Far & Near & Far & Near & Far & Near & Far\\
        \midrule
        \multirow{2}{*}{ResNet18/50} & EBO & 87.58 & 91.21 & \textbf{80.91} & 79.77 & 82.50 & \textbf{90.86} & 75.89 & 89.47 \\
        & BHL & \textbf{90.25} & \textbf{96.00} & 71.57 & \textbf{86.08} & \textbf{86.72} & 76.13 & \textbf{79.04} & \textbf{89.75}\\
        \midrule
        \multirow{2}{*}{EfficientNet-B7} & EBO & \textbf{97.39} & 98.91 & \textbf{87.46} & 86.91 & 75.02 & 86.53 & 65.16 & 81.65 \\
        & BHL & 87.43 & \textbf{99.74} & 84.21 & \textbf{99.80} & \textbf{78.83} & \textbf{93.02} & \textbf{85.24} & \textbf{94.49} \\
        \midrule
        \multirow{2}{*}{ViT-B-16} & EBO & \textbf{90.91} & 93.9 & \textbf{88.81} & 87.23 & \textbf{69.72} & \textbf{83.49} & 62.93 & 78.71 \\
        & BHL & 79.38 & \textbf{96.14} & 81.38 & \textbf{97.98} & 62.19 & 81.40 & \textbf{74.06} & \textbf{88.43} \\
        
        \bottomrule
    \end{tabular}
    \caption{EBO and BHL compared on different models.}
    \label{tab:otherarchs}
    \end{table*}
}

{
    \newcommand{\tablefullold}[1]{
        \begin{table*}[h]
        \caption{}
        \footnotesize
        \label{tab:main_results}
        \centering
        \begin{tabular}{l|ccc|ccc|ccc|ccc} 
            \toprule
             & \multicolumn{3}{c|}{\textbf{CIFAR-10}} & \multicolumn{3}{c|}{\textbf{CIFAR-100}} & \multicolumn{3}{c|}{\textbf{ImageNet-200}} & \multicolumn{3}{c}{\textbf{ImageNet-1K}} \\
            
             & Near & Far & ID Acc. & Near & Far & ID Acc. & Near & Far & ID Acc. & Near & Far & ID Acc. \\
            \midrule
        
        \rowcolor{NO_TRAIN} MSP & 88.03 & 90.73 & 95.06 & 80.27 & 77.76 & 77.25 & 83.34 & 90.13 & 86.37 & 76.02 & 85.23 & 76.18 \\
        
        \rowcolor{NO_TRAIN} TempScale & 88.09 & 90.97 & 95.06 & 80.90 & 78.74 & 77.25 & \textbf{83.69} & 90.82 & 86.37 & 77.14 & 87.56 & 76.18 \\
        
        \rowcolor{NO_TRAIN} ODIN & 82.87 & 87.96 & 95.06 & 79.90 & 79.28 & 77.25 & 80.27 & 91.71 & 86.37 & 74.75 & 89.47 & 76.18 \\
        
        \rowcolor{NO_TRAIN} Gram & 58.66 & 71.73 & 95.06 & 51.66 & 73.36 & 77.25 & 67.67 & 71.19 & 86.37 & 61.70 & 79.71 & 76.18 \\
        
        \rowcolor{NO_TRAIN} EBO & 87.58 & 91.21 & 95.06 & 80.91 & 79.77 & 77.25 & 82.50 & 90.86 & 86.37 & 75.89 & 89.47 & 76.18 \\
        
        \rowcolor{NO_TRAIN} GradNorm & 54.90 & 57.55 & 95.06 & 70.13 & 69.14 & 77.25 & 72.75 & 84.26 & 86.37 & 72.96 & 90.25 & 76.18 \\
        
        \rowcolor{NO_TRAIN} ReAct & 87.11 & 90.42 & 95.06 & 80.77 & 80.39 & 77.25 & 81.87 & 92.31 & 86.37 & 77.38 & 93.67 & 76.18 \\
        
        \rowcolor{NO_TRAIN} MLS & 87.52 & 91.10 & 95.06 & \textbf{81.05} & 79.67 & 77.25 & 82.90 & 91.11 & 86.37 & 76.46 & 89.57 & 76.18 \\
        
        \rowcolor{NO_TRAIN} KLM & 79.19 & 82.68 & 95.06 & 76.56 & 76.24 & 77.25 & 80.76 & 88.53 & 86.37 & 76.64 & 87.60 & 76.18 \\
        
        \rowcolor{NO_TRAIN} VIM & 88.68 & \textbf{93.48} & 95.06 & 74.98 & 81.70 & 77.25 & 78.68 & 91.26 & 86.37 & 72.08 & 92.68 & 76.18 \\
        
        \rowcolor{NO_TRAIN} KNN & \textbf{90.64} & 92.96 & 95.06 & 80.18 & 82.40 & 77.25 & 81.57 & 93.16 & 86.37 & 71.10 & 90.18 & 76.18 \\
        
        \rowcolor{NO_TRAIN} DICE & 78.34 & 84.23 & 95.06 & 79.38 & 80.01 & 77.25 & 81.78 & 90.80 & 86.37 & 73.07 & 90.95 & 76.18 \\
        
        \rowcolor{NO_TRAIN} ASH & 75.27 & 78.49 & 95.06 & 78.20 & 80.58 & 77.25 & 82.38 & \textbf{93.90} & 86.37 & \textbf{78.17} & \textbf{95.74} & 76.18 \\
        
        \rowcolor{NO_TRAIN} SHE & 81.54 & 85.32 & 95.06 & 78.95 & 76.92 & 77.25 & 80.18 & 89.81 & 86.37 & 73.78 & 90.92 & 76.18 \\
        
        \multicolumn{7}{l}{\textbf{- Training Methods (w/ Outlier Data)}} \vspace{.1cm} \\
        
        \rowcolor{EXTRA_DATA} OE & \textbf{94.82} & \textbf{96.00} & 94.63 & \textbf{88.30} & \textbf{81.41} & \textbf{76.84} & \textbf{84.84} & \textbf{89.02} & 85.82 & \textcolor{gray}{N/A} & \textcolor{gray}{N/A} & \textcolor{gray}{N/A} \\
        
        \rowcolor{EXTRA_DATA} MCD & 91.03 & 91.00 & 94.95 & 77.07 & 74.72 & 75.83 & 83.62 & 88.94 & \textbf{86.12} & \textcolor{gray}{N/A} & \textcolor{gray}{N/A} & \textcolor{gray}{N/A} \\
        
        \rowcolor{EXTRA_DATA} UDG & 89.91 & 94.06 & 92.36 & 78.02 & 79.59 & 71.54 & 74.30 & 82.09 & 68.11 & \textcolor{gray}{N/A} & \textcolor{gray}{N/A} & \textcolor{gray}{N/A} \\
        
        \rowcolor{EXTRA_DATA} MixOE & 88.73 & 91.93 & 94.55 & 80.95 & 76.40 & 75.13 & 82.62 & 88.27 & 85.71 & \textcolor{gray}{N/A} & \textcolor{gray}{N/A} & \textcolor{gray}{N/A} \\
        
        \bottomrule
        \end{tabular}
        \end{table*}
    }

    \newcommand{\tableaggregation}[1]{
        \begin{table*}[h]
            \centering
            \small
            \begin{tabular}{l|cc|cc|cc|cc}
                \toprule
                &\multicolumn{2}{c|}{\textbf{CIFAR-10}}&\multicolumn{2}{c|}{\textbf{CIFAR-100}}&\multicolumn{2}{c|}{\textbf{ImageNet200}}&\multicolumn{2}{c}{\textbf{ImageNet}}\\
                & \textbf{Near OoD} & \textbf{Far OoD} & \textbf{Near OoD} & \textbf{Far OoD} & \textbf{Near OoD} & \textbf{Far OoD} & \textbf{Near OoD} & \textbf{Far OoD} \\
                \midrule
                \textbf{EBO} & 87.65 & 91.04 &79.71&79.85&\textbf{90.36}&\textbf{81.42}&76.16&87.56 \\
                \midrule
                \textbf{BHL} & \textbf{90.25} & 96.00 &71.57&86.08&86.72&76.13&\textbf{79.04}&\textbf{89.75}\\
                \midrule
                \textbf{BHL w/ Reg.} & 87.64 & \textbf{96.9} &69.64&\textbf{90.13}&N/A&N/A&N/A&N/A\\
                \midrule
                \textbf{Aggr. w/ KNN} & 85.09 & 88.98 &60.69&59.59&57.42&68.25&53.03&58.29 \\
                \textbf{Aggr. w/ Mahalanobis} & 66.53 &87.59&57.09&80.01&57.16&77.32&59.46&78.93 \\
                \textbf{Aggr. w/ VAE} &81.77&86.24&68.18&69.12&58.49&63.97&51.36&57.1 \\
                \bottomrule
            \end{tabular}
            \caption{.}
            \label{tab:aggregation}
        \end{table*}
    }

    \newcommand{\tablecifar}[1]{
        \begin{table*}[h]
            \centering
            \small
            \setlength{\tabcolsep}{6pt}
            \renewcommand{\arraystretch}{1.2}
            \begin{tabular}{l|ccc|ccccc}
                \toprule
                & \textbf{CIFAR-100} & \textbf{TIN} & \textbf{Near OoD} & \textbf{MNIST} & \textbf{Places365} & \textbf{SVHN} & \textbf{Texture} & \textbf{Far OoD} \\
                \midrule
                \textbf{EBO} & 86.46 & 88.85 & 87.65 & 94.32 & 88.45 & 91.98 & 89.40 & 91.04 \\
                \textbf{BHL} & \textbf{88.23} & \textbf{92.26} & \textbf{90.25} & \textbf{99.89} & \textbf{92.13} & \textbf{98.46} & \textbf{93.5} & \textbf{96.00} \\
                \bottomrule
            \end{tabular}
            \caption{AUROC scores of EBO, BHL and three aggregation methods with \textsc{CIFAR-10} and \textsc{CIFAR-100} as $\mathcal{D}_{in}$, averaged over 3 runs.}
            \label{tab:cifar10}
        \end{table*}
    }

    \newcommand{\tablecifarhund}[1]{
        \begin{table*}[h]
            \centering
            \small
            \setlength{\tabcolsep}{6pt}
            \renewcommand{\arraystretch}{1.2}
            \begin{tabular}{l|ccc|ccccc}
                \toprule
                & \textbf{CIFAR-10} & \textbf{TIN} & \textbf{Near OoD} & \textbf{MNIST} & \textbf{Places365} & \textbf{SVHN} & \textbf{Texture} & \textbf{Far OoD} \\
                \midrule
                \textbf{EBO} & \textbf{79.00} & \textbf{80.43} & \textbf{79.71} & 79.18 & \textbf{79.51} & 82.30 & 78.41 & 79.85 \\
                \textbf{BHL} & 72.54 & 70.60 & 71.57 & \textbf{99.94} & 68.53 & \textbf{98.78} & \textbf{80.90} & \textbf{86.08} \\
                \bottomrule
            \end{tabular}
            \caption{AUROC scores of EBO and BHL with \textsc{CIFAR-100} as $\mathcal{D}_{in}$, averaged over 3 runs.}
            \label{tab:cifar100}
        \end{table*}
    }

    \newcommand{\tableimagenettwo}[1]{
        \begin{table*}[h]
            \centering
            \small
            \setlength{\tabcolsep}{6pt}
            \renewcommand{\arraystretch}{1.2}
            \begin{tabular}{l|ccc|cccc}
                \toprule
                & \textbf{SSB-Hard} & \textbf{NINCO} & \textbf{Near OoD} & \textbf{iNaturalist} & \textbf{Texture} & \textbf{OpenImage-O} & \textbf{Far OoD} \\
                \midrule
                \textbf{EBO} & \textbf{78.44} & \textbf{84.39} & \textbf{81.42} & \textbf{92.67} & \textbf{89.69} & \textbf{88.69} & \textbf{90.36} \\
                \textbf{BHL}   & 74.36 & 77.9 & 76.13 & 90.12 & 87.66 & 82.4 & 86.72 \\
                \bottomrule
            \end{tabular}
            \caption{AUROC scores of EBO and BHL with \textsc{ImageNet200} ($\mathcal{D}_{in}$), averaged over 3 runs.}
            \label{tab:imagenet200}
        \end{table*}
    }
    
    \newcommand{\tableimagenet}[1]{
        \begin{table*}[h]
            \centering
            \small
            \setlength{\tabcolsep}{6pt}
            \renewcommand{\arraystretch}{1.2}
            \begin{tabular}{l|ccc|cccc}
                \toprule
                & \textbf{SSB-Hard} & \textbf{NINCO} & \textbf{Near OoD} & \textbf{iNaturalist} & \textbf{OpenImage-O} & \textbf{Texture} & \textbf{Far OoD} \\
                \midrule
                \textbf{EBO} & 71.94 & \textbf{80.39} & 76.16 & 89.21 & \textbf{88.00} & 85.47 & 87.56 \\
                \textbf{BHL} & \textbf{79.72} & 78.36 & \textbf{79.04} & \textbf{95.69} & 78.94 & \textbf{94.63} & \textbf{89.75} \\
                \bottomrule
            \end{tabular}
            \caption{AUROC scores of EBO and BHL with \textsc{ImageNet} as $\mathcal{D}_{in}$, averaged over 1 run.}
            \label{tab:imagenet}
        \end{table*}
    }
}